\documentclass{egpubl}
\usepackage{eg2024}

\usepackage[T1]{fontenc}
\usepackage{dfadobe}  

\usepackage{cite}  
\BibtexOrBiblatex

\usepackage{times}
\usepackage{epsfig}
\usepackage{graphicx}
\usepackage{amsmath}
\usepackage{amssymb}
\usepackage{booktabs}
\usepackage{pifont}
\usepackage{multirow}
\usepackage{makecell}
\usepackage{xcolor}
\usepackage{ifsym}
\usepackage{etoolbox}
\usepackage{xspace}
\usepackage{dcolumn}

\usepackage{soul}
\usepackage{amsmath}
\usepackage{amssymb}
\usepackage{microtype}
\usepackage{wrapfig}
\usepackage{fancyhdr}
\usepackage{mathtools}
\usepackage{xspace}
\usepackage{float}

\def\figurePath{figures/}
\def\myfigure#1#2#3{\begin{figure}[htb]\centering\includegraphics*[width = #3\columnwidth]{\figurePath#1}\\[-1.5ex]\caption{#2}\vspace{-0.5cm}\label{fig:#1}\end{figure}}

\def\mycfigure#1#2{\begin{figure*}[htb]\centering\includegraphics*[clip, width = \textwidth]{\figurePath#1}\\[-1.5ex]\caption{#2}\vspace{-0.5cm}\label{fig:#1}\end{figure*}}

\def\myhfigure#1#2{\begin{figure*}[htb!]\centering\includegraphics*[clip, width = 0.9\textwidth]{\figurePath#1}\\[-2.0ex]\caption{#2}\vspace{-0.5cm}\label{fig:#1}\end{figure*}}

\newcommand{\refSec}[1]{Sec.~\ref{sec:#1}}
\newcommand{\refFig}[1]{Fig.~\ref{fig:#1}}

\newcommand{\refTbl}[1]{Tbl.~\ref{tbl:#1}}

\newcommand{\FLIP}{FLIP}

\soulregister\ref7
\soulregister\cite7
\soulregister\refFig7
\soulregister\refSec7
\soulregister\ref7
\soulregister\pageref7
\soulregister\shortcite7
\soulregister\eg0
\soulregister\ie0

\DeclareGraphicsExtensions{.png,.jpg,.pdf,.ai,.psd}
\newcommand{\mysubsection}[2]{\subsection{#1}\label{sec:#2}}

\newcommand{\myparagraph}[1]{\vspace{0.15cm} \noindent\textbf{#1}\ \ }

\newcommand{\colorIcon}[2]{\textcolor{color#1}{\csname icon#2\endcsname}}

\newcommand{\nameAndIcon}[1]{``#1'' ({\csname icon#1\endcsname})}
\newcommand{\nameAndIconA}[1]{(``#1'', {\csname icon#1\endcsname})}

\definecolor{red}{rgb}{0.9,0,0}

\usepackage[pagebackref=true,breaklinks=true,colorlinks,bookmarks=false]{hyperref}

\newcommand{\mymath}[2]{\newcommand{#1}{\TextOrMath{$#2$\xspace}{#2}}}

\mymath{\reference}{X}
\mymath{\distorted}{Y}
\mymath{\referenceFeature}{X_f}
\mymath{\distortedFeature}{Y_f}
\mymath{\visualmask}{M}
\mymath{\IQM}{\mathcal D}
\mymath{\EnhancedIQM}{\bar \IQM}
\mymath{\maskgenerator}{\mathcal F}
\mymath{\numberofchannels}{\pmb{\mathtt{F}}}
\mymath{\mos}{q}
\mymath{\scalernetwork}{\mathcal G}
\mymath{\Loss}{Loss}

\usepackage{colortbl}

\usepackage{enumitem}

\usepackage{arydshln} 
\title{Enhancing image quality prediction with self-supervised visual masking}

\author[\c{C}o\u{g}alan et al.]
{\parbox{\textwidth}{\centering U\u{g}ur \c{C}o\u{g}alan, Mojtaba Bemana, Hans-Peter Seidel and Karol Myszkowski
	}
	\\
	{\parbox{\textwidth}{\centering Max-Planck-Institut f\" ur Informatik, Germany
		}
	}
}

\begin{document}

\definecolor{fixedcolor}{rgb}{1,.95,.4}

\newcommand{\mojtaba}[1]{#1}
\DeclareRobustCommand{\change}[1]{{\sethlcolor{fixedcolor}\hl{#1}}}

\teaser{
   \vspace{-1.0cm}
   \includegraphics[width=\textwidth]{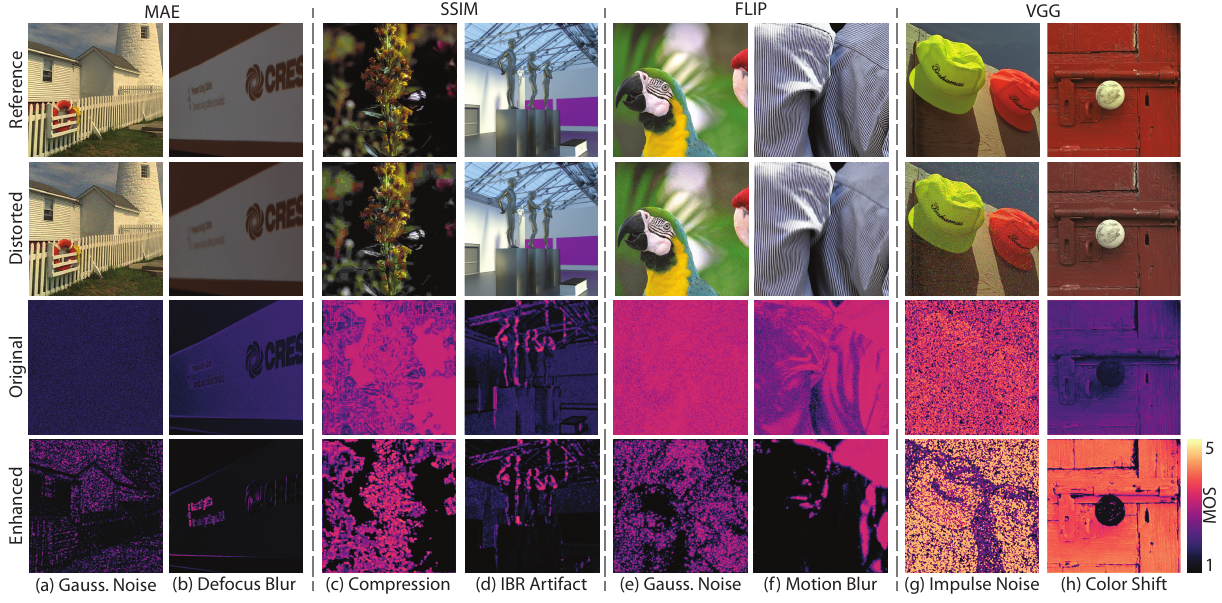}
   \centering
   \caption{
    We introduce self-supervised visual masking that enhances image quality prediction for existing quality metrics such as MAE, SSIM, \FLIP, and VGG. Our work is inspired by the well-known characteristic of the Human Visual System (HVS), visual masking, which results in locally varying sensitivity to image artifact visibility that reduces with increasing contrast magnitude of the original image pattern. We found that the learned masking clearly outperforms its traditional hand-crafted versions and better adapts to specific distortion patterns. In the first two rows, we show the reference and distorted images, while the third and fourth rows show the error maps as predicted by the original metrics and their enhanced versions using our masking approach. As can be seen, our mask-enhanced metrics better predict the local distortion visibility by the human observer.  For a more intuitive comparison, we scale each error map to fit within the mean opinion scores (MOS) range (please refer to \refSec{evals} for more details). In this color scale, darker indicates less visible distortion.}
   \label{fig:Teaser}
   }
   
\maketitle

\begin{abstract}
Full-reference image quality metrics (FR-IQMs) aim to measure the visual differences between a pair of reference and distorted images, with the goal of accurately predicting human judgments. However, existing FR-IQMs, including traditional ones like PSNR and SSIM and even perceptual ones such as HDR-VDP, LPIPS, and DISTS, still fall short in capturing the complexities and nuances of human perception.
In this work, rather than devising a novel IQM model, we seek to improve upon the perceptual quality of existing FR-IQM methods. We achieve this by considering visual masking, an important characteristic of the human visual system that changes its sensitivity to distortions as a function of local image content. Specifically, for a given FR-IQM metric, we propose to predict a visual masking model that modulates reference and distorted images in a way that penalizes the visual errors based on their visibility.  Since the ground truth visual masks are difficult to obtain, we demonstrate how they can be derived in a self-supervised manner solely based on mean opinion scores (MOS) collected from an FR-IQM dataset. Our approach results in enhanced FR-IQM metrics that are more in line with human prediction both visually and quantitatively. Plase refer to \url{https://enhancediqm.mpi-inf.mpg.de/} for supplementary materials.


\end{abstract}





\myhfigure{2afc_results}{Agreement of metric predictions with human judgments. We consider the classic (MAE and SSIM) and learning-based (LPIPS and DISTS) metrics, and we compare their prediction to their enhanced versions (E-MAE, E-SSIM, E-DISTS, and E-LPIPS) using our approach. On the left, we see a situation where MAE and SSIM favor JPEG-like artifacts over slightly resampled textures. On the right, we encounter a scenario where LPIPS and DISTS prefer blur over a subtle color shift. Our extended metric versions are better aligned with human choice. The images have been extracted from the PIPAL dataset \cite{gu2020pipal}. }

\section{Introduction}
Full-Reference Image Quality Metrics (FR-IQMs), which take as an input a pair of reference and distorted images, play a crucial role in a wide range of applications in digital image processing, such as image compression and transmission, as well as in evaluating the rendered content in computer graphics and vision. They are commonly used as a cost function in optimizing restoration tasks like denoising, deblurring, and super-resolution \cite{ding2021comparison}. 
Consequently, developing FR-IQMs that accurately reflect the visual quality of images in accordance with the characteristics of the human visual system (HVS) is critical. The most commonly used FR-IQMs for evaluating image quality are the mean square error (MSE) or mean absolute error (MAE). While these per-pixel metrics are easy to compute, they assess image quality regardless of spatial content, leading to false positive predictions. 
This can be seen in \refFig{Teaser}a, where Gaussian noise is less noticeable in textured regions, while MAE predicts uniformly distributed error. Similarly, a depth-of-field blur is primarily visible on high-contrast fonts \refFig{Teaser}b, while MAE predicts the blur visibility also in smooth gradient regions.
Other classic metrics like SSIM \cite{wang04}, while accounting for spatial content, often result in false positive predictions (the JPEG artifact and image-based rendering (IBR) artifact in \refFig{Teaser}c-d, respectively). 
A recent hand-crafted metric \FLIP \,  \cite{Andersson2020} is specifically designed to predict the visual differences in time-sequential image-pair flipping, which can make it too sensitive for side-by-side image evaluation, e.g., noise is less visible in high-contrast texture (\refFig{Teaser}e) or \mojtaba{motion blur is not equally visible across different parts of an image} (\refFig{Teaser}f).
Recognizing that hand-crafted image features may not adequately capture the HVS complexity, modern metrics \cite{zhang2018perceptual} strive to assess the perceptual dissimilarity between images by comparing deep features extracted from classification networks \cite{simonyan2015deep}. These metrics appear to better account for the HVS characteristics; however, they are designed to generate a single value per image pair and cannot provide correct visible error localization, as can be seen in the impulse noise example (\refFig{Teaser}g).
Moreover, the features learned through training the classification networks tend to be less sensitive to global distortions, such as moderate color and brightness changes (\refFig{Teaser}h) that have less impact on reliable classification.

The objective of this work is not to develop a new perceptual FR-IQM; instead, we are interested in improving the quality prediction of existing metrics to align more closely with human judgment (\refFig{2afc_results}). We also aim to enhance the accuracy of error map predictions by considering multiple factors such as image content, distortion levels, and distortion types. By detecting both the presence and evaluating the magnitude of visible distortion in each pixel, we aim to ensure that the metric predictions more accurately reflect the probability of a human observer detecting differences between a pair of images. In this regard, there have been several efforts toward incorporating the perceptual aspects of human vision, specifically visual masking \cite{legge1980field,foley94human,wilson1984modified}, into FR-IQM methods \cite{lubin1995visual,Daly93,mantiuk2011hdr,mantiuk2021fovvideovdp}. In simple words, visual masking refers to the phenomenon in which certain components of an image (in our application, distortions) may be less visible to the viewer due to the presence of other visual elements in the same image. Visual masking can affect image quality perception, making some image distortions less visible to the viewer \cite{ferwerda1997model,zeng2020overview}. However, existing visual masking models are typically hand-crafted and struggle to generalize effectively across various distortion types. 
 Although learning a visual masking model appears to be a natural solution, the lack of reliable ground truth data for visual masking makes direct supervision impractical. In this work, we propose a self-supervised approach to predict visual masking using a dataset of images featuring a variety of distortions of different magnitudes whose quality has been evaluated in the mean opinion scores (MOS) experiment with human subjects \cite{lin19}. 
In summary, our work offers the following contributions: 
\begin{itemize}
\item We propose a lightweight CNN that generates a mask for a given reference and distorted input pair. The predicted mask acts as a per-pixel weight and, when multiplied with the inputs, greatly improves the performance of the existing FR-IQMs. The incorporation of our learned mask into any FR-IQM is seamless and demands minimal computational resources. While the CNN is trained specifically for each metric, it learns a generic masking model capable of identifying various types of distortions.
\item We demonstrate that our masking model can be generalized to deep features and used as a per-layer feature map weight. 
\item Our solution significantly enhances the accuracy of quality prediction for FR-IQMs across various test datasets. Furthermore, it produces per-pixel error maps that visually align more closely with human perception compared to the original FR-IQMs.
\item We show the potential application of our approach as a loss function for training image denoising and motion deblurring.
\end{itemize}
 \mycfigure{overview}{Our proposed visual masking for enhancing classic metrics such as MAE and SSIM (left) and learning-based metrics such as DISTS or LPIPS (right). For classic metrics, the input to our mask predictor network \maskgenerator are sRGB images, while for learning-based metrics, the inputs are the VGG features extracted from the images. We learn the visual masks in a self-supervised fashion by minimizing the difference between the metric final score and human scores collected from an FR-IQM dataset.}

\section{Previous work}

FR-IQMs can be categorized into classical metrics, which perform the computation directly in the image space, and learning-based metrics, which leverage deep feature models to assess image quality. 

\myparagraph{Classic metrics}
Basic FR-IQMs, such as MSE, RMSE, and MAE,
compute the per-pixel difference to quantify image distortion. While these metrics are straightforward to calculate, their consistency with human vision is typically low. Such perceptual consistency can be improved by considering relative instead of absolute error, as in PSNR and the symmetric mean absolute percentage error (SMAPE) \cite{vogels2018denoising}.
To account for the spatial aspects of the HVS, alternative metrics such as SSIM \cite{wang04} are introduced, which consider image patches and measure local differences in luminance, contrast, and structural information. SSIM is further extended to multi-scale MS-SSIM \cite{wang03} and complex wavelet CW-SSIM \cite{sampat2009complex} versions that capture both global and local structural information. FSIM \cite{zhang11} decomposes the image into multiple subbands using Gabor filters and compares subband responses between the reference and distorted images. 
By assuming that natural images have a specific distribution of pixel values, models based on information theory \cite{sheikh2005information,sheikh06} measure the mutual information between images by comparing their joint histograms and taking into account the statistical dependencies between neighboring pixels. Classical metrics can offer either a single overall quality score or a visibility map indicating the distortion intensity. Watson-DCT \cite{Watson1993}, VDM \cite{lubin1995visual}, VDP \cite{Daly93}, HDR-VDP \cite{mantiuk2011hdr}, and fovVideoVDP \cite{mantiuk2021fovvideovdp} measure either the visibility of distortions or perceived distortions magnitude, or both by considering various visual aspects such as luminance adaptation, contrast sensitivity, and visual masking. A more recent metric, \FLIP \, \cite{Andersson2020}, emphasizes color differences, and it is sensitive to even subtle distortions by emulating flipping between the compared image pair.

\myparagraph{Deep learning-based metrics}
In recent years, research in FR-IQM has been placing greater emphasis on perceptual comparisons in deep feature space rather than image space to enhance the alignment with human judgments. Prashnani et al. \cite{prashnani2018pieapp} are among the first to utilize deep feature models learned from human-labeled data to predict perceptual errors. Zhang et al.  \cite{zhang2018perceptual} demonstrate that internal image representations from classification networks can be used for image comparison. They propose the Perceptual Image Patch Similarity (LPIPS) index, which quantifies image similarity by measuring the $\ell_2$ distances between pre-trained VGG features. To further improve the correlation with human judgments, they learn per-channel weights for selected VGG features using their collected perceptual similarity dataset. Recognizing that simple $\ell_p$-norm measures fail to consider the statistical dependency of errors across different locations, Ding et al. \cite{ding20} introduce the DISTS, which aims to measure the texture and structure similarity between feature pairs by comparing their global mean, variance, and correlations in the form of SSIM. Building upon this work, A-DISTS \cite{ding21} extended the approach to incorporate local structure and texture comparisons. Czolbe et al. \cite{Czolbe2020} incorporate their extended Watson-DCT model \cite{Watson1993} as a measure of VGG feature distance. Moving away from deterministic point-wise feature comparisons, DeepWSD \cite{liao22} compares the overall distributions of features using the Wasserstein distance, a statistical measure for comparing two distributions. Nevertheless, the majority of the proposed IQMs metrics are targeted toward producing a single quality score and are not primarily designed to generate per-pixel error maps. In this regard, Wolski et al. \cite{wolski2018dataset} employ a custom CNN model trained in a fully supervised way using coarse user marking data to predict an error visibility map that highlights the regions where distortions are more likely to be noticeable. \\
Recently, deep learning-based no-reference metrics (NR-IQM) such as KonCept512 \cite{koniq10k}, HYPERIQA \cite{hyperiqa20}, MUSIQ \cite{musiq21} 
and MANIQA \cite{yang2022maniqa} have been proposed. While NR-IQM methods often report impressive performance, their practical applicability remains limited. FR-IQM metrics are still predominant in CG applications, as the reference images are typically readily available.


In this work, we extend the classic and deep learning-based full-reference metrics by introducing a learnable component trained on perceptual MOS data in a self-supervised way. By implicitly analyzing local image content, our model derives per-pixel maps that mimic visual masking, effectively modeling the visual significance of distortions.

\section{Self-supervised visual masking}

This section elaborates on our methodology for perceptually calibrating the existing FR-IQMs. Given a reference and distorted pair (\reference and \distorted) $ \in R^{H \times W \times C}$, we first learn a visual mask, \visualmask $ \in R^{H \times W \times 1}$, which has the same spatial dimensions as the inputs.  For classical metrics (\refFig{overview}-left), the input \reference and \distorted are sRGB images ($C = 3$), while for learning-based metrics such as LPIPS, DISTS, or DeepWSD, the input \reference and \distorted are the VGG features extracted from the images and $C$ is the number of channels in a given VGG layer (\refFig{overview}-right). 
The predicted mask is then element-wise multiplied with  \reference and \distorted before being fed into an FR-IQM, \IQM. Note that, for learning-based metrics, a direct modulation of the input sRGB images by a mask \visualmask would distort their content and consequently reduce the VGG performance as it is originally trained on complete, non-masked images.
Our solution with VGG feature modulation draws inspiration from classic FR-IQMs  \cite{lubin1995visual,Daly93,mantiuk2011hdr,mantiuk2021fovvideovdp}, where the response from hand-crafted filter banks is transduced using a fixed, perception-motivated masking model \cite{legge1980field,foley94human,wilson1984modified}. In our approach, the response from pre-trained VGG filters is modulated with a learned per-pixel mask \visualmask, where perception modeling is learned from the MOS data.
We estimate the mask \visualmask by utilizing a lightweight CNN denoted as \maskgenerator, which takes both \reference and \distorted as input. Mathematically, this can be expressed as:
\begin{align}
\visualmask = \maskgenerator (\reference ,  \distorted)
\end{align}

It is important to note that the network \maskgenerator is trained specifically for a metric \IQM. In the case of metrics such as LPIPS, DISTS, and DeepWSD, we follow their specific architecture and compute a mask for each layer using a separate \maskgenerator, and the same mask is applied for all channels in a given layer (\refFig{overview}-right). The original spatial pooling is preserved for each metric, such as L1 distance in LPIPS, structural similarity in DISTS, or Wasserstein distance in DeepWSD. Since we cannot directly supervise the output of the mask generator network, we adopt a self-supervised approach to train it using an IQM dataset with a single quality score. The network's parameters are optimized by minimizing the $ \ell_{2}$ difference between the metric output value and human scores. Our loss is formulated as follows:
\begin{align}
\Loss = \lVert  \scalernetwork(\IQM(\visualmask \odot \reference, \visualmask \odot \distorted)) - \mos \rVert_{2}^{2}
\end{align}

Here, \mos $\in [0,1]$ represents the normalized mean opinion score when comparing the images \reference and \distorted. As the metric response can vary in an arbitrary range, following a similar approach in \cite{zhang2018perceptual}, a small network  \scalernetwork is jointly trained to map the metric response to the human ratings. 

\subsection{Training and network details}
For training, we use the KADID dataset \cite{lin19}, which comprises 81 natural images that have been distorted using 25 types of traditional distortions, each at five different levels, making roughly 10k training pairs. Note that we train our mask generator network \maskgenerator for all the distortion categories together rather than for one specific category. We find that a lightweight CNN with three convolutional layers, each consisting of 64 channels, suffices for successful training. ReLU activation is applied after each layer, while we use Sigmoid activation for the final layer to keep the mask values in the range between 0 and 1. The computation overhead of our mask generator network is very negligible, and it takes only 12\,ms to compute the mask on a 1080Ti GPU with an input resolution of $768 \times 512 \times 3$. Our mapping network \scalernetwork consists of two 32-channel fully connected (FC) ReLU layers, followed by a 1-channel FC layer with Sigmoid activation. The batch size for training is set to 4. We employ the Adam optimizer \cite{kingma2017adam} with an initial learning rate of $10^{-4}$ and a weight decay of $10^{-6}$. 

\mycfigure{figure_visibility_metrics}{
Visual comparisons of distortion visibility maps for Gaussian noise (upper row) and superresolution artifacts (middle and bottom rows). The distortion examples are taken from the PIPAL dataset. The first two columns present the reference and distorted images, followed by the respective metric predictions: MAE, HDR-VDP-2 \cite{mantiuk2011hdr}, LocVis \cite{wolski2018dataset}, FovVideoVDP \cite{mantiuk2021fovvideovdp}, and our E-MAE. Here, we additionally visualize the MAE map to better understand the characteristics of each distortion. As can be seen, the existing metrics tend to either underestimate or overestimate the distortion visibility. Note that LocVis and E-MAE have not seen distorted images with superresolution artifacts in their training.}

\definecolor{lightgray}{RGB}{230, 230, 230}
\definecolor{darkgray}{RGB}{205, 205, 205}

 \vspace{-0.2cm}
 \begin{table}[h!]
     \centering
     \setlength{\tabcolsep}{2pt}
     \renewcommand{\arraystretch}{1.15}
     \caption{Performance comparison of existing FR-IQMs and their enhanced versions using our approach (specified by the prefix E) on three standard IQM datasets. The prefix R denotes the original metric retraining on the KADID dataset, while the prefix S refers to employing a visual saliency mask instead of our mask. At the bottom, we include the corresponding results for NR-IQMs. Higher values of SRCC, PLCC, and KRCC indicate better quality prediction. The first and second best metrics for each dataset are indicated in bold and underlined, respectively.  Additionally, the version with superior correlation is highlighted in dark gray for each metric. }
     \resizebox{\columnwidth}{!}{%
     \begin{tabular}{lc cccccccccc}
         \toprule
         &
         \multicolumn3c{\textbf{CSIQ}}&
         \multicolumn3c{\textbf{TID}} &
         \multicolumn3c{\textbf{PIPAL}} & 
         \\
         \cmidrule(lr){2-4}
         \cmidrule(lr){5-7}
         \cmidrule(lr){8-10}
         \multicolumn1l{\textbf{Metric}}&
         \multicolumn1c{PLCC}&
         \multicolumn1c{SRCC}&
         \multicolumn1c{KRCC}&
         \multicolumn1c{PLCC}&
         \multicolumn1c{SRCC}&
         \multicolumn1c{KRCC}&
         \multicolumn1c{PLCC}&
         \multicolumn1c{SRCC}&
         \multicolumn1c{KRCC}&\\
       \midrule
    

FSIM & 0.900 & 0.913 & 0.740 & 0.847 & 0.789 & 0.611  & 0.651 & 0.617 & 0.441 \\
VIF & 0.826 & 0.841 & 0.642 & 0.820 & 0.813 & 0.616  & 0.584 & 0.538 & 0.378 \\
HDR-VDP-2 & 0.761 & 0.886 & 0.704 & 0.715 & 0.753 & 0.571 & 0.514 & 0.503 & 0.354 \\
PieAPP & 0.827 & 0.840 & 0.653 & 0.832 & 0.849 & 0.652 & \textbf{0.729} & \textbf{0.709} & \textbf{0.521} \\
\midrule

MAE &  \cellcolor{lightgray} 0.819 & \cellcolor{lightgray} 0.801 \cellcolor{lightgray} & \cellcolor{lightgray} 0.599 & \cellcolor{lightgray} 0.639 & \cellcolor{lightgray} 0.627 &\cellcolor{lightgray}  0.409 & \cellcolor{lightgray} 0.458 & \cellcolor{lightgray} 0.443 & \cellcolor{lightgray} 0.304 \\
S-MAE &  \cellcolor{lightgray} 0.656 & \cellcolor{lightgray} 0.697 \cellcolor{lightgray} & \cellcolor{lightgray} 0.493 & \cellcolor{lightgray} 0.498 & \cellcolor{lightgray} 0.496 & \cellcolor{lightgray}  0.347 & \cellcolor{lightgray} 0.369 & \cellcolor{lightgray} 0.365 & \cellcolor{lightgray} 0.248 \\
E-MAE & \cellcolor{darkgray} 0.871 & \cellcolor{darkgray} 0.917 & \cellcolor{darkgray} 0.738 & \cellcolor{darkgray} 0.857 & \cellcolor{darkgray} 0.863 & \cellcolor{darkgray} 0.673  & \cellcolor{darkgray} 0.597 & \cellcolor{darkgray} 0.606 & \cellcolor{darkgray} 0.429 \\
\hdashline

PSNR & \cellcolor{lightgray} 0.851 & \cellcolor{lightgray} 0.837 \cellcolor{lightgray} & \cellcolor{lightgray} 0.645 & \cellcolor{lightgray} 0.726 & \cellcolor{lightgray} 0.714 & \cellcolor{lightgray} 0.540  & \cellcolor{lightgray} 0.468 & \cellcolor{lightgray} 0.456 & \cellcolor{lightgray} 0.314 \\
E-PSNR & \cellcolor{darkgray} 0.901 & \cellcolor{darkgray} 0.910 & \cellcolor{darkgray} 0.728 &\cellcolor{darkgray}  0.855 & \cellcolor{darkgray} 0.844 & \cellcolor{darkgray} 0.656  & \cellcolor{darkgray} 0.637 & \cellcolor{darkgray} 0.629 & \cellcolor{darkgray} 0.446 \\
\hdashline 

SSIM & \cellcolor{lightgray} 0.848 & \cellcolor{lightgray} 0.863 &\cellcolor{lightgray} 0.665 &\cellcolor{lightgray} 0.697 &\cellcolor{lightgray} 0.663 &\cellcolor{lightgray} 0.479  &\cellcolor{lightgray} 0.550 &\cellcolor{lightgray} 0.534 &\cellcolor{lightgray} 0.373 \\
E-SSIM &  \cellcolor{darkgray} 0.869 &\cellcolor{darkgray} 0.910 &\cellcolor{darkgray} 0.732 &\cellcolor{darkgray} 0.842 & \cellcolor{darkgray} 0.868 & \cellcolor{darkgray}  0.677 & \cellcolor{darkgray}  0.671 & \cellcolor{darkgray}  0.656 & \cellcolor{darkgray}  0.469 \\
\hdashline 

MS-SSIM & \cellcolor{lightgray} 0.826 & \cellcolor{lightgray} 0.841 & \cellcolor{lightgray} 0.642 & \cellcolor{darkgray} 0.820 & \cellcolor{lightgray} 0.813 & \cellcolor{lightgray} 0.616 & \cellcolor{lightgray} 0.584 & \cellcolor{lightgray} 0.538 & \cellcolor{lightgray} 0.379 \\
E-MS-SSIM &\cellcolor{darkgray} 0.862 &\cellcolor{darkgray} 0.895 &\cellcolor{darkgray} 0.709 &\cellcolor{lightgray} 0.806 &\cellcolor{darkgray} 0.825 &\cellcolor{darkgray} 0.621 &\cellcolor{darkgray} 0.642 &\cellcolor{darkgray} 0.634 &\cellcolor{darkgray} 0.453 \\
\hdashline 

\FLIP &  \cellcolor{lightgray}0.731 & \cellcolor{lightgray} 0.724 & \cellcolor{lightgray} 0.527 & \cellcolor{lightgray} 0.591 & \cellcolor{lightgray} 0.537 & \cellcolor{lightgray} 0.413 & \cellcolor{lightgray} 0.498 &  \cellcolor{lightgray} 0.442 & \cellcolor{lightgray} 0.306\\
E-FLIP &\cellcolor{darkgray} 0.871 &\cellcolor{darkgray} 0.902 &\cellcolor{darkgray} 0.715 &\cellcolor{darkgray} 0.859 &\cellcolor{darkgray} 0.858 &\cellcolor{darkgray} 0.666 &\cellcolor{darkgray} 0.621 &\cellcolor{darkgray} 0.612 & \cellcolor{darkgray} 0.434 \\
\hdashline 

FovVideoVDP & \cellcolor{lightgray} 0.795  & \cellcolor{lightgray} 0.821 & \cellcolor{lightgray} 0.632 & \cellcolor{lightgray} 0.742 & \cellcolor{lightgray} 0.727 & \cellcolor{lightgray} 0.544 & \cellcolor{lightgray} 0.565 & \cellcolor{lightgray} 0.509 & \cellcolor{lightgray} 0.358 \\
E-FovVideoVDP & \cellcolor{darkgray} 0.841 & \cellcolor{darkgray} 0.882 & \cellcolor{darkgray} 0.685 & \cellcolor{darkgray} 0.830 & \cellcolor{darkgray} 0.816 & \cellcolor{darkgray} 0.623  & \cellcolor{darkgray} 0.662 & \cellcolor{darkgray} 0.626 & \cellcolor{darkgray} 0.449 \\
\hdashline 

VGG & \cellcolor{darkgray} 0.938 & \cellcolor{darkgray} \underline{0.952} & \cellcolor{darkgray} 0.804 & \cellcolor{lightgray} 0.853 & \cellcolor{lightgray} 0.820 & \cellcolor{lightgray} 0.639 & \cellcolor{lightgray} 0.643 & \cellcolor{lightgray} 0.610 & \cellcolor{lightgray} 0.432\\
E-VGG & \cellcolor{lightgray} 0.914 & \cellcolor{lightgray} 0.938 & \cellcolor{lightgray} 0.776 & \cellcolor{darkgray} \underline{0.895} & \cellcolor{darkgray} \underline{0.889} & \cellcolor{darkgray} \underline{0.710}  & \cellcolor{darkgray} 0.695 & \cellcolor{darkgray} 0.675 & \cellcolor{darkgray} 0.485 \\
\hdashline 

LPIPS & \cellcolor{darkgray} 0.944 & \cellcolor{lightgray} 0.929 & \cellcolor{lightgray} 0.769 & \cellcolor{lightgray} 0.803 & \cellcolor{lightgray} 0.756 & \cellcolor{lightgray} 0.568  & \cellcolor{lightgray} 0.640 & \cellcolor{lightgray} 0.598 & \cellcolor{lightgray} 0.424 \\
R-LPIPS & \cellcolor{lightgray} 0.931 & \cellcolor{lightgray} 0.917 & \cellcolor{lightgray} 0.756 & \cellcolor{darkgray} 0.898 & \cellcolor{darkgray} 0.886 & \cellcolor{darkgray} 0.697  & \cellcolor{lightgray} 0.670 & \cellcolor{lightgray} 0.640 & \cellcolor{lightgray} 0.447 \\
E-LPIPS & \cellcolor{lightgray} 0.922 & \cellcolor{darkgray} 0.933 & \cellcolor{darkgray} 0.771 & \cellcolor{lightgray} 0.884 & \cellcolor{lightgray} 0.876 & \cellcolor{lightgray} 0.689  & \cellcolor{darkgray} 0.705 & \cellcolor{darkgray} 0.678 & \cellcolor{darkgray} 0.490 \\
\hdashline 

DISTS & \cellcolor{darkgray} 0.947 & \cellcolor{darkgray} 0.947 & \cellcolor{darkgray} 0.796 & \cellcolor{lightgray} 0.839 & \cellcolor{lightgray} 0.811 & \cellcolor{lightgray} 0.619 & \cellcolor{lightgray} 0.645 & \cellcolor{lightgray} 0.626 & \cellcolor{lightgray} 0.445 \\
E-DISTS & \cellcolor{lightgray} 0.938 & \cellcolor{lightgray} 0.925 & \cellcolor{lightgray} 0.754 & \cellcolor{darkgray} \textbf{0.903} & \cellcolor{darkgray} \textbf{0.915} & \cellcolor{darkgray} \textbf{0.725}  & \cellcolor{darkgray} \underline{0.725} & \cellcolor{darkgray} \underline{0.697} & \cellcolor{darkgray} \underline{0.507} \\
\hdashline 

Watson-VGG & \cellcolor{darkgray} 0.944 & \cellcolor{darkgray} 0.940 & \cellcolor{darkgray} 0.785 & \cellcolor{lightgray} 0.808 & \cellcolor{lightgray} 0.763 & \cellcolor{lightgray} 0.573 & \cellcolor{lightgray} 0.627 & \cellcolor{lightgray} 0.606 & \cellcolor{lightgray} 0.429 \\
E-Watson-VGG & \cellcolor{lightgray} 0.917 & \cellcolor{lightgray} 0.936 & \cellcolor{lightgray} 0.776 & \cellcolor{darkgray} 0.886 & \cellcolor{darkgray} 0.895 & \cellcolor{darkgray} 0.716  & \cellcolor{darkgray} 0.697 & \cellcolor{darkgray} 0.678 & \cellcolor{darkgray} 0.488 \\
\hdashline 

DeepWSD & \cellcolor{lightgray} \underline{0.949} & \cellcolor{darkgray} \textbf{0.961} & \cellcolor{lightgray} \underline{0.821} & \cellcolor{lightgray} 0.879 & \cellcolor{lightgray} 0.861 & \cellcolor{lightgray} 0.674 & \cellcolor{lightgray} 0.593 & \cellcolor{lightgray} 0.584 & \cellcolor{lightgray} 0.409 \\
R-DeepWSD & \cellcolor{darkgray} \textbf{0.955} & \cellcolor{darkgray} \textbf{0.961} & \cellcolor{darkgray} \textbf{0.823} & \cellcolor{lightgray} 0.895 & \cellcolor{lightgray} 0.88 & \cellcolor{lightgray} 0.695 & \cellcolor{lightgray} 0.654 & \cellcolor{lightgray} 0.633 & \cellcolor{lightgray} 0.449 \\
E-DeepWSD & \cellcolor{lightgray}0.937 & \cellcolor{lightgray} 0.937 & \cellcolor{lightgray} 0.775 & \cellcolor{darkgray} 0.905 & \cellcolor{darkgray} 0.892 & \cellcolor{darkgray} 0.710 & \cellcolor{darkgray} 0.704 & \cellcolor{darkgray} 0.672 & \cellcolor{darkgray} 0.485\\

\midrule
\midrule
HYPERIQA & 0.769 & 0.757 & 0.573 & 0.679 & 0.662 & 0.489 & 0.325 & 0.363 & 0.250 \\
MANIQA & 0.874 & 0.827 & 0.642 & 0.784 & 0.760 & 0.572 & 0.404 & 0.407 & 0.276 \\


     \bottomrule 
     \end{tabular}}
     \label{tbl:numbers}
     \vspace{-0.3cm}
 \end{table}
 \raggedbottom

\section{Results}
In this section, we first present our experimental setup, which we use for our method evaluation and ablations of different training strategies. 

\subsection{Experimental setup}
We employ our visual masking approach to enhance some of the classical metrics (MAE, PSNR, SSIM, MS-SSIM, \FLIP, and fovVideoVDP) and recent learning-based methods (VGG, LPIPS, DISTS, Watson-VGG, and DeepWSD). Note for MS-SSIM, we used the same \maskgenerator across all scales, while the inputs are images at different scales. Moreover, the metric called VGG is computed by simply taking the $\ell_{1}$ difference between VGG features for the same layers as originally chosen for LPIPS and DISTS. Deploying our masking model to PieAPP or any other metrics that create new CNN architectures from scratch is not practical as there is no intermediate component to which we can apply our masking model. Thus, our main focus remains on mainstream metrics that use features extracted from pre-trained networks for quality prediction. We assess the performance of our proposed approach on three well-established IQM datasets: CSIQ \cite{Larson2010MostAD}, TID2013 \cite{Ponomarenko13}, and PIPAL \cite{gu2020pipal}. The first two datasets mainly consist of synthetic distortions, ranging from 1k to 3k images. On the other hand, PIPAL is the most comprehensive IQM dataset due to its diverse and complex distortions, consisting of 23k images. Each reference image in this dataset was subjected to 116 distortions, including 19 GAN-type distortions. For evaluation, following \cite{ding20}, we resize the smaller side resolution of input images to 224 while maintaining the aspect ratio. Note that rescaling is only performed on the test datasets to match the image resolution in which the MOS data were collected. Our approach does not require rescaled inputs, and all visual figures in the paper are processed in their original resolution.
For each dataset, three metrics are used for evaluation: Spearman's rank correlation coefficient (SRCC), Pearson linear correlation coefficient (PLCC), and the Kendall rank correlation coefficient (KRCC). The PLCC measures the accuracy of the predictions, while the SRCC indicates the monotonicity of the predictions, and the KRCC measures the ordinal association. 
The PLCC measures linear correlation, requiring both variables (metric output and MOS) to be on the same scale, hence, we mapped the metric scores to the MOS values using a four-parameter logistic function, consistent with established IQM methods \cite{ding20,liao22}. We do not use \scalernetwork for PLCC remapping; otherwise, we need to train a specific \scalernetwork for each metric on a given test set. Importantly, SRCC and KRCC scores do not require additional remapping, thus directly reflecting the correlation between metric output and MOS data.

\mysubsection{Evaluations}{evals}
In this section, we present the outcome of the quantitative (agreement with the MOS data) and qualitative (the quality of error maps) evaluation of our method. We also analyze the mask content and relate it with perceptual models of contrast and blur perception. Finally, we analyze the error map prediction of different distortion levels, and we consider the potential use of our enhanced E-MAE metric as a loss in denoising and deblurring image restoration tasks.

\myparagraph{Quality prediction}
The experimental results are presented in \refTbl{numbers}, where with the prefix E, we denote our proposed extension for each specific IQM. 
Our extension of traditional metrics, such as MAE, PSNR, SSIM, \FLIP, and fovVideoVDP, consistently improves their performance for all datasets. This is remarkable as those metrics are commonly used, and our simple extension can make their distortion prediction closer to the human observer. Interestingly, the enhanced E-MAE and E-PSNR outperform recent learning-based VGG, LPIPS, and DISTS in the TID dataset while showing a comparable performance for the PIPAL dataset. Notable improvements are also observed in both datasets for the recent learning-based metrics (E-VGG, E-LPIPS, E-DIST, Watson-VGG, and E-DeepWSD), positioning them at a level comparable to other state-of-the-art IQMs, such as PieAPP  \cite{prashnani2018pieapp}. The only exception is the case of the small-scale CSIQ dataset, where the original learning-based metrics achieve high correlations with the MOS data and leave little space for further improvements. Please see our supplementary for a more detailed analysis.

We also consider retraining LPIPS per-channel weights using the KADID dataset (denoted as R-LPIPS in \refTbl{numbers}), which improves correlation for TID and PIPAL datasets with respect to the original LPIPS. 
Compared to our E-LPIPS, such retraining is more prone to overfitting; it performs marginally better for the TID dataset, which has more distortion similarities with KADID, while it is significantly worse for the larger and more diverse PIPAL dataset.
Similarly, training layer-specific weights for DeepWSD (R-DeepWSD) improves correlation, however, our E-DeepWSD achieves better performance. 
 Moreover, channel/layer-wise weighing can not be reasonably applied to image-based metrics (MAE, SSIM, FLIP).

We evaluate the performance of recent NR-IQM methods MANIQA \cite{yang2022maniqa} and HYPERIQA \cite{hyperiqa20} that are trained on the KADID dataset. As it can be seen in \refTbl{numbers}, the NR-IQM methods show significantly lower correlations with the MOS data, particularly for the PIPAL dataset, which indicates that FR-IQM methods can better generalize to unseen distortion types.

Visual saliency methods incorporate semantic information, however, they are not trained to discriminate between dominant distortions and salient features (e.g., faces). This seems to be a limiting factor in the direct saliency use in our image quality evaluation framework. To validate this observation, we employed a predicted saliency map from an off-the-shelf saliency network \cite{jia18} as a mask to the MAE metric that we denote as S-MAE in \refTbl{numbers}. While in this simple attempt, we observe significantly lower correlations with the MOS data, we believe that our approach can be complemented by saliency, so that effectively distortion predictions are narrowed to image regions that are likely to be visually attended.


\myparagraph{Error map prediction}
In \refFig{Teaser}, we show the error maps predicted by various existing IQMs and their enhanced versions for a set of images featuring different types of distortions. 
As the output of each metric can be in an unbounded range and vary across different metrics and their improved versions with our approach, for a more intuitive and fair comparison, instead of simply normalizing them within the range from zero to one using a Sigmoid function \cite{Andersson2020}, we visualize the output of each metric after being scaled to the MOS range using a pre-trained scaling network {\scalernetwork}. Specifically, we utilize KADID dataset and train a separate {\scalernetwork}  for each metric to transform their raw response into values that align with human ratings (MOS). Note that for the enhanced version of each metric, the network {\scalernetwork} is already provided from the training step. In general, this scaling process is akin to mapping the metric scores to the MOS values using a four-parameter function when computing the correlation.
As can be seen in \refFig{Teaser}, the enhanced error maps using our approach better align with the human perception of the distortion. A notable example is the case of Gaussian noise, where a metric like MAE predicts uniformly distributed error, and our masking approach effectively redistributes the error in terms of both their magnitude and local visibility. We provide more examples in our supplementary materials. Additionally, \refFig{figure_visibility_metrics} showcases three examples where our E-MAE metric achieves \mojtaba{ better localized} error maps compared to well-established visibility metrics such as HDR-VDP-2 \cite{mantiuk2011hdr}, LocVis \cite{wolski2018dataset}, and FovVideoVDP \cite{mantiuk2021fovvideovdp}. 

\myfigure{mask_noise_blur}{Comparison of our E-MAE metric masks for the noise (fifth row) and blur (sixth row) distortions as a function of different image contrast ($\times 0.5, \times 1,$ and $\times 2$). In the fourth row, we also show a map with the human sensitivity to local contrast changes as predicted by a traditional model of visual contrast masking \cite[Eq.4]{tursun2019luminance}. In all cases, darker means more masking (less sensitive to distortion).}{1.0}



\myparagraph{Mask visualization}
It is also intriguing to see the learned mask, i.e., the output of the network \maskgenerator, and to compare it with a traditional visual contrast masking model, such as the one used in JPEG2000 compression \cite{zeng2020overview}. To this end, \refFig{mask_noise_blur} presents our masks generated for noise and blur distortions. We consider the same distortion level and three levels of image contrast enhancement ($\times 0.5, \times 1,$ and $\times 2$). 
In the case of noise distortion, our learned masks predict stronger visual masking in the high-contrast butterfly and better noise visibility in the out-of-focus smooth background.  Increasing image contrast ($\times 2$) leads to even stronger visual masking in the butterfly area and the plant behind it.  Reducing image contrast ($\times 0.5$) leads to the inverse effect. Such behavior is compatible with the visual contrast masking model \cite{zeng2020overview,tursun2019luminance}, where due to self-contrast masking, the higher the contrast of the original signal (e.g., on edges), the stronger the distortion should be to make it visible. Along a similar line, due to neighborhood masking, the higher the contrast texture, the stronger the visual masking as well.
In the case of blur distortion, our learned mask predicts its strong visibility on high-contrast edges. The stronger the image contrast ($\times 2$), the blur visibility improves. Assigning a higher weight by our mask to high contrast regions agrees with perceptual models of blur detection and discrimination \cite{watson2011blur,stephen2015defocus}. \\
Note that we derive each mask taking as an input both the reference and distorted images; the mask can resolve even per-pixel distortions, as in the case of noise (\refFig{mask_noise_blur}), and accordingly informs the E-MAE metric on the perceptual importance of such distortions.
What is also remarkable is that the HVS might impose contradictory requirements on hand-crafted visual models that become specific for a given distortion. This is well illustrated in \refFig{mask_noise_blur}, where noise can be better masked by strong contrast patterns \cite{zeng2020overview,tursun2019luminance} while blur is actually better revealed by strong contrast patterns \cite{watson2011blur}. Our learned E-MAE mask recognizes the distortion context and reacts as expected by penalizing less noise distortion in high-contrast and textured regions while penalizing more blur distortion at high-contrast edges. Interestingly, such local, seemingly contradictory behavior has been learned solemnly by providing multiple pairs of reference and distortion images along with the corresponding quality MOS rating, which is just a single number. 
No annotation on specific distortion types has been required in our training.
\refFig{perceptual_pattern} shows further examples that our learned masking is also informed about contrast masking by texture \cite{ferwerda1997model} and the contrast sensitivity function (CSF) \cite{Daly93,Barten:1999,Mantiuk2020}.


\myparagraph{Masks vs. metrics analysis}
Masks typically vary with distortion type, as demonstrated in \refFig{mask_noise_blur} for noise and blur. In \refFig{mask_visualisation}, we further illustrate the predicted mask across various metrics, including MAE, PSNR, SSIM, FLIP, and VGG for a given pair of reference and distorted images with Gaussian noise.
As can be seen, metrics with similar characteristics, such as MAE, PSNR, and FLIP, tend to learn similar maps.  For a more perceptually-informed metric like SSIM that partially models visual masking, our predicted mask adjusts its sensitivity by assigning lower weight to regions where SSIM exaggerates the error (e.g., in the grass area) and identity weight when accurately predicting the error magnitude (e.g., the body of lighthouse). \mojtaba{When it comes to VGG, the mask learned for the early layer resembles the MAE mask since the initial convolutional layers tend to learn basic image features like edges and textures, whereas for the deeper layers, as the VGG learns more abstract features, the interpretation of the masks become less obvious. In the supplementary, we provide an additional example with blur distortion. }

\mycfigure{mask_visualisation}{Visualisation of predicted mask across different metrics for a given pair of reference and distorted images with Gaussian noise from the TID dataset. Note that the SSIM values have been remapped to 1-SSIM, where lower values indicate less visible errors. \mojtaba{ In the case of the PSNR, we show the error map for the measured MSE}. For the VGG metric, we visualize the predicted mask for all layers, while the error map is shown only for the first layer. \vspace{0.3cm} }


\mycfigure{app}{Visual results in the image denoising task when employing MAE and MAE+E-MAE as loss functions. Considering that denoiser networks typically reduce noise through smoothing, our objective was to investigate whether the use of the E-MAE loss component could encourage the network to retain or hallucinate details, even if they do not precisely match the reference but their discrepancy from the ground truth is possibly not perceivable. As can be seen, the denoised images with the MAE+E-MAE loss yield sharper content and higher contrast.}

\myparagraph{Employing the enhanced metric as a loss}
In this part, we investigate the benefit of the enhanced IQMs in optimizing image restoration algorithms. To this end, we employ MAE and MAE+$\lambda \cdot$ E-MAE as loss functions for training image denoising and motion deblurring using the state-of-the-art image restoration method, Restormer \cite{zamir2022restormer}. 
For the denoising task, we select the images in the BSD400 dataset \cite{MartinFTM01} as our training set and introduce synthetic noise to these images by applying additive white Gaussian noise with a randomly chosen standard deviation ranging between 0 and 50. We performed each training with the same number of iterations in an identical setup (e.g., learning rate).
Then, we evaluate the trained models on five benchmark datasets, consistent with the ones used in \cite{zamir2022restormer}. We conduct our evaluation for various noise levels and report the results in \refTbl{number_loss}. 
We can observe that training just with the MAE loss leads to higher PSNRs, in particular for higher noise levels, but at the same time, image blur and contrast loss can be observed (refer to  \refFig{app}). More perceptually inclined quality metrics penalize for such visual quality reduction, e.g., LPIPS is sensitive to excessive blur \cite{zhang2018perceptual}. Combining with an E-MAE loss component clearly improves such metrics' scores consistently across various noise levels as well as the visual quality.  
For the motion deblurring task, we employed the GoPro dataset \cite{nah2017deep} for the training and evaluation. The combination of MAE and E-MAE enhances the deblurring results across different quality metrics ({\refTbl{number_loss_deblurring}}) and leads to a sharper appearance ({\refFig{app_deblurring}}). 
In both tasks, we empirically found that $\lambda=1$ gives the best performance.  We also observe that relying exclusively on the E-MAE loss component leads to worse results, which is expected, as indicated in \cite{ding2021comparison}. In the supplementary, we provide more comparisons for other loss combinations, such as MAE+VGG and MAE+E-VGG that lead to similar conclusions.

 \begin{table*}[]
     \setlength{\tabcolsep}{4pt}
     \caption{Evaluation of a blind Gaussian denoising task when employing MAE and the equal combination of MAE and E-MAE as loss functions. We show the performance of the trained models on synthetic Gaussian noise created with four distinct noise levels ($\sigma$) averaged across five benchmark datasets, consistent with the ones used in \cite{zamir2022restormer}. }
     \resizebox{\textwidth}{!}{%
     \begin{tabular}{lc cccccccccccccccc}
          \toprule
        &
        \multicolumn4c{$\sigma = 15$}&
        \multicolumn4c{$\sigma = 25$}&
        \multicolumn4c{$\sigma = 50$}&
        \multicolumn4c{$\sigma = 60$}
        \\
        \cmidrule(lr){2-5}
        \cmidrule(lr){6-9}
        \cmidrule(lr){10-13}
        \cmidrule(lr){14-17}
        \multicolumn1l{\textbf{Loss}}&
        \multicolumn1c{PSNR$\uparrow$}&
        \multicolumn1c{SSIM$\uparrow$}&
        \multicolumn1c{LPIPS$\downarrow$}&
        \multicolumn1c{E-MAE$\downarrow$}&
        \multicolumn1c{PSNR$\uparrow$}&
        \multicolumn1c{SSIM$\uparrow$}&
        \multicolumn1c{LPIPS$\downarrow$}&
        \multicolumn1c{E-MAE$\downarrow$}&
        \multicolumn1c{PSNR$\uparrow$}&
        \multicolumn1c{SSIM$\uparrow$}&
        \multicolumn1c{LPIPS$\downarrow$}&
        \multicolumn1c{E-MAE$\downarrow$}&
        \multicolumn1c{PSNR$\uparrow$}&
        \multicolumn1c{SSIM$\uparrow$}&
        \multicolumn1c{LPIPS$\downarrow$}&
        \multicolumn1c{E-MAE$\downarrow$}
        \\
        \midrule  
    
MAE & 34.36 & \textbf{0.94}& 0.058 & 0.0343 & \textbf{31.94}& 0.90 &0.092&0.0849 & \textbf{28.82}&\textbf{0.84}&0.163&3.187 &  \textbf{28.02}&\textbf{0.81}&0.182&4.258 \\

MAE + E-MAE &  \textbf{34.37} & \textbf{0.94}& \textbf{0.055} &\textbf{0.0145} & 31.92 & \textbf{0.91}&\textbf{0.087}&\textbf{0.0263} & 28.71& \textbf{0.84} &\textbf{0.152} &\textbf{0.790} & 
27.88 & \textbf{0.81}&\textbf{0.167}&\textbf{1.035} \\


     \bottomrule
     \end{tabular}}
     
     \label{tbl:number_loss}
 \end{table*}
 \raggedbottom


\myhfigure{app_deblurring}{Visual results for the motion deblurring task when employing MAE and MAE + E-MAE as loss functions. \vspace{0.2cm}}

\subsection{Ablations}
We perform a set of ablations to investigate the impact of reduced training data in terms of distortion levels, reference image number, and distortion type diversity on the E-MAE metric prediction accuracy.

\myparagraph{Distortion levels} The first experiment analyzes the importance of incorporating various distortion levels into our training set. In this regard, we train our network for the E-MAE metric using only one distortion level per category, and the results are reported in \refFig{level}. Interestingly, for all the datasets (except PIPAL), an inverse U-shape trend emerged across five different distortion levels, where we observe the lowest correlation when training with the minimum and maximum distortion levels (levels 1 and 5). Conversely, a moderate amount of distortion (level 3) appears to be sufficiently representative for each distortion category and achieved a comparable correlation to training with all five levels.
This behavior can be anticipated because, at the lowest and highest distortion levels, the distortions are either barely visible or strongly visible, leading to the consistent selection of mostly extreme rating scores.
Consequently, when the network is exclusively exposed to images with one such extreme distortion and rating levels, it fails to learn to differentiate between them. On the other hand, at moderate distortion levels where distortions are partially visible or invisible, the network has a better opportunity to learn masks that behave differently for varying spatial locations.

\begin{table}[]
     \centering
    \setlength{\tabcolsep}{3pt}
    \caption{
    Evaluation of a motion deblurring task when employing MAE and the equal combination of MAE and E-MAE as loss functions. We show the performance of the trained models on synthetic blur created using the GoPro dataset \cite{nah2017deep}.} 
    \resizebox{6.0cm}{!}{
    \begin{tabular}{lc cccc}
        \toprule
        \multicolumn1l{\textbf{Metric} \hspace{1cm}}& 
        \multicolumn1c{PSNR$\uparrow$}&
        \multicolumn1c{SSIM$\uparrow$}&
        \multicolumn1c{LPIPS$\downarrow$}&
        \multicolumn1c{E-MAE$\downarrow$}&
       \\
       \midrule

MAE & 31.70 & 0.92 & 0.1030 & 0.0192 \\
MAE + E-MAE & \textbf{31.78} & \textbf{0.93}& \textbf{0.1018} & \textbf{0.0184} \\

    \bottomrule
    \end{tabular}}
    \vspace{-0.3cm}
   \label{tbl:number_loss_deblurring}
\end{table}
\raggedbottom

\myfigure{level}{Evaluation of E-MAE training performance using only selected distortion levels for each distortion category. We measure the SRCC correlation with the MOS data, and as a reference, we also include the results of complete training with all distortion levels. 
}{0.8}

\myparagraph{Dataset size} Although we employ a large-scale  KADID dataset in our training (25 distortion types $\times$ five distortion levels), the number of reference images is limited to 81.
This ablation aims to investigate the training performance by even further reducing the number of reference images. 
To this end, we perform multiple runs of E-MAE metric training using randomly selected subsets of 20, 40, and 60 reference images.
\refFig{number} presents the SRCC correlations averaged over multiple runs. 
The correlation differences between 40, 60, and the full set of 81 reference images are minor. In the case of 20 reference images, the performance is slightly lower and the variance higher, which indicates that 20 scenes might not be enough to capture image content variability.

\myfigure{number}{Evaluation of E-MAE training performance using different numbers of the reference images (scenes). 
Multiple training runs have been performed for 20, 40, and 60 randomly selected scenes from the full set of 81 reference images. 
The data points represent the respective SRCC correlation averages over such runs, while the vertical bars depict the standard deviation.}{0.8}



\myparagraph{Distortion diversity} We investigate the impact of separate E-MAE training on specific distortion subsets such as noise, blur, combined noise, and blur, as well as the complete KADID dataset. At the test time, we evaluate trained this way E-MAE versions on noise and blur subsets of the TID dataset, as well as its complete version.
The results, presented in \refTbl{percategorytraining}, reveal that training solely on the noise category unsurprisingly improves the SRCC correlation within that category; however, it also enhances the overall correlation for the TID dataset with respect to the original MAE. Conversely, training exclusively on blur does not improve the performance within the blur category itself, as the blur distortion already exhibits a strong correlation (0.934) for the MAE metric, making any improvement marginal. On the other hand, we noticed that training with all categories combined significantly improves the correlation in both the noise and blur categories compared to training with only noise or blur categories, which can suggest that exposing the network to a wider range of distortion types enables better generalization. 

In our supplementary material, we additionally show the impact of each of the three factors on the predicted error maps.

\begin{table}[]
    \centering
    \setlength{\tabcolsep}{3pt}
    \caption{
    The SRCC correlation with the MOS data for the E-MAE metric trained with specific distortion categories (noise, blur, noise\&blur) and the entire (all) KADID dataset, as indicated in the brackets. The TID dataset is used for testing, where the ``Category'' columns indicate whether only the noise and blur subsets are considered or the entire dataset. 
    }
    \resizebox{6.0cm}{!}{
    \begin{tabular}{lc ccccc}
        \toprule
        \multicolumn1l{\textbf{Metric} \hspace{1cm} \small \textbf{Category:}}& 
       \multicolumn1c{noise}&
       \multicolumn1c{blur} &
       \multicolumn1c{all}
       \\
       \midrule
    MAE &  0.601 & 0.934 & 0.545 \\
    E-MAE (noise) & 0.847 & 0.927 & 0.674 \\
    E-MAE (blur) & 0.732 & 0.926 & 0.655 \\
    E-MAE (noise \& blur) & 0.841 & 0.936 & 0.726 \\
    E-MAE (all) & \textbf{0.906} & \textbf{0.955} & \textbf{0.857} \\    
    \bottomrule
    \end{tabular}}
    \vspace{-0.3cm}
   \label{tbl:percategorytraining}
\end{table}
\raggedbottom

\mycfigure{perceptual_pattern}{Error map prediction for the MAE and E-MAE metrics along with learned weighting masks for two perception patterns from \cite{vcadik2013learning}. These patterns were specifically designed to investigate various perceptual phenomena, including contrast sensitivity and contrast masking. In the first row, the background consists of a high-frequency pattern with increasing contrast toward the right and a stimulus pattern with decreasing contrast from bottom to top (which becomes more apparent when zoomed in). In this scenario, contrast masking is more pronounced with increasing background contrast that, in turn, reduces the stimulus visibility, and E-MAE correctly predicts this effect. 
The second row presents another example, showing a set of patterns where their spatial frequencies increase toward the right while their contrast decreases toward the top. 
In this case, the learned masking roughly follows an inverse U-shape characteristic, akin to the contrast sensitivity function (CSF) \cite{Daly93,Barten:1999,Mantiuk2020}.
Our masking well approximates the sensitivity drop for high frequencies but tends to suppress the visibility of low-frequency patterns excessively.
In spite of this drawback, we still find it quite remarkable that the CSF shape becomes apparent in our learned mask without any explicit training with calibrated near-threshold CSF data.\vspace{-0.2cm}}

\section{Limitations and future work}
The actual visual contrast masking is the function of the viewing condition and the display size \cite{chandler2013seven},
which is often considered in the perceptual quality metrics \cite{Daly93,mantiuk2011hdr,mantiuk2021fovvideovdp,Andersson2020} but otherwise mostly ignored. 
However, the effectiveness of our visual masking model is limited to the experimental setup where human scores are obtained in the KADID dataset. 

As we have demonstrated, incorporating our masking model into traditional metrics is straightforward, but it might be a difficult task for certain network architectures, such as PieAPP \cite{prashnani2018pieapp}.

As shown in \refFig{perceptual_pattern}, in the context of the CSF reproduction, our metric might not be well calibrated for near contrast threshold stimuli, whose visibility is also affected by the viewing distance and display conditions. 
Wolski et al. \cite{wolski2018dataset} developed the LocVis dataset with local maps of distortion detection probability that emphasize near-threshold distortions. Unfortunately, the LocVis dataset is not reliable for supra-threshold distortion in terms of their magnitude estimation, while we readily learn from the MOS data. We relegate as future work, combining such not-compatible distortion visibility and magnitude estimation data into a consistent training dataset for enhancing our masking model.




\section{Conclusion}

In this paper, we present a new approach towards reducing the notorious gap between the existing quality metric prediction and the actual distortion visibility by the human observer.
We achieve this by self-supervised training of a metric-specific network using the existing distortion datasets labeled with mean opinion score (MOS). 
We show that although overall image quality is rated with a single MOS value in the training data, by securing sufficient diversity of such training, as detailed in our ablation study, the network can leverage global MOS into a meaningful per-pixel mask.
The mask, through different weighting of local distortion visibility, which also adapts to specific distortion types, helps a given metric to aggregate such local information into the comprehensive MOS value, as imposed by the training data.
The mask can be learned directly in the image space for traditional metrics or in the feature space for recent learning-based metrics. In either case, it is trivial to incorporate into most of the existing metrics.
Remarkably, our approach improves the performance of commonly used metrics, such as MAE, PSNR, SSIM, and \FLIP\, on all datasets we tested. The prediction accuracy of recent learning-based metrics is typically substantially enhanced. 

\bibliographystyle{eg-alpha}
\bibliography{paper}


\end{document}